\documentclass[a4paper, fleqn]{cas-dc}
\usepackage{amsmath,amssymb,amsfonts}
\usepackage{algorithmic}
\usepackage{graphicx}
\usepackage{textcomp}
\usepackage{algorithmic}
\usepackage{algorithm}
\usepackage{array}
\usepackage{afterpage}
\usepackage{multirow}       
\usepackage[justification=centering]{caption} 
\usepackage{threeparttable} 
\usepackage[caption=false,font=normalsize,labelfont=sf,textfont=sf]{subfig}
\usepackage{stfloats}
\usepackage{url}
\usepackage{verbatim}
\usepackage[switch, pagewise]{lineno}  
\usepackage[authoryear]{natbib}
\usepackage{hyperref}

\def\tsc#1{\csdef{#1}{\textsc{\lowercase{#1}}\xspace}}
\tsc{WGM}
\tsc{QE}

\begin{document}
\let\WriteBookmarks\relax
\def\floatpagepagefraction{1}
\def\textpagefraction{.001}
\shorttitle{SDDA for Cross-Session MI Classification}    

\shortauthors{Miao et al.}  

\title[mode = title]{Priming Cross-Session Motor Imagery Classification with A Universal Deep Domain Adaptation Framework}

\author[1]{Zhengqing Miao}
\ead{mzq@tju.edu.cn}

\author[2]{Xin Zhang}
\ead{xin_zhang_bme@tju.edu.cn}

\author[3]{Carlo Menon}
\ead{carlo.menon@hest.ethz.ch}

\author[1]{Yelong Zheng}
\ead{zhengyelongby@tju.edu.cn}

\author[1]{Meirong Zhao}
\fnmark[*]
\ead{MeirongZhao@tju.edu.cn}

\author[2]{Dong Ming}
\ead{richardming@tju.edu.cn}

\affiliation[1]{organization={Laboratory of Optoelectronic Detection and Image Processing, Department of Instruments Science and Technology, School of Precision Instrument and Opto-electronics Engineering, Tianjin University},
            city={Tianjin},
            postcode={300072}, 
            country={China}}
            
\affiliation[2]{organization={Laboratory of Neural Engineering and Rehabilitation, Department of Biomedical Engineering, School of Precision Instruments and Optoelectronics Engineering, Tianjin University, and also with the Tianjin International Joint Research Center for Neural Engineering, Academy of Medical Engineering and Translational Medicine, Tianjin University},
            city={Tianjin},
            postcode={300072}, 
            country={China}}
            
\affiliation[3]{organization={Biomedical and Mobile Health Technology Laboratory, ETH Zurich},
            city={Lengghalde 5, 8008 Zurich},
            country={Switzerland}}


\cortext[1]{Corresponding author: M. Zhao}

\begin{keywords}
Domain Adaptation \sep Deap Learning \sep  Brain Computer Interface (BCI) \sep Electroencephalogram (EEG) \sep Motor Imagery (MI)
\end{keywords}
\maketitle

\begin{abstract}
Electroencephalogram (EEG) based motor imagery (MI) brain computer interfaces (BCI) are widely used in applications related to rehabilitation and external device control. However,  due to the non-stationary and low signal-to-noise ratio characteristics of EEG, classifying motor imagery tasks of the same participant from different recording sessions is generally challenging. Whether the classification accuracy of cross-session MI can be improved from the perspective of domain adaptation is a question worth verifying. 
In this paper, we propose a Siamese deep domain adaptation (SDDA) framework for cross-session MI classification based on mathematical models in domain adaptation theory. The SDDA framework primarily consists of three components: a novel preprocessing method based on domain-invariant features, a maximum mean discrepancy (MMD) loss for aligning source and target domain embedding features, and an improved cosine-based center loss designed to suppress the influence of noise and outliers on the neural network. The SDDA framework has been validated with two classic and popular convolutional neural networks (EEGNet and ConvNet) from BCI research field in two MI EEG public datasets (BCI Competition IV IIA, IIB). Compared with the vanilla EEGNet and ConvNet, the SDDA framework improves the MI classification accuracy by 10.49\%, 7.60\% respectively in IIA dataset, and 4.59\%, 3.35\% in IIB dataset. The SDDA not only significantly improves the classification performance of the vanilla networks but also surpasses state-of-the-art transfer learning methods, making it a superior and user-friendly approach for MI classification. 
\end{abstract}

\section{Introduction}
Brain-computer interfaces (BCIs) are communication systems between the brain and external devices that do not rely on peripheral nervous and muscular systems (\cite{wolpaw2007brain}). BCIs are usually used to assist, enhance, repair, or partially replace human cognitive or sensorimotor functions (\cite{wolpaw2000brain}). Motor imagery (MI) is a widely used paradigm in EEG-based BCIs. Results from neuroscience studies suggest that MI induces event-related synchronization/desynchronizations (ERD/ERS) in the Mu band, which are the key discriminant features that are used in MI-BCIs (\cite{Pfurtscheller2006}). Feature extraction algorithms, such as common spatial pattern (CSP) (\cite{pfurtscheller2001motor}) and filter bank common spatial pattern (FBCSP) (\cite{ang2008filter}) have been successfully used in conjunction with classifiers, such as linear discriminant analysis (LDA) or support vector machine (SVM), in the literature for MI-BCIs applications.

With the development of artificial neural network (ANN) in computer vision and natural language processing, ANNs and their applications in the research field of BCIs are attracting more and more attention (\cite{schwemmer2018meeting}), \cite{al2021deep}, \cite{craik2019deep}).
Compared with traditional decoding methods that manually extract features, ANNs are able to extract complex features of EEG without prior feature assumption. For example, ConvNet (\cite{schirrmeister2017deep}) has played a significant role in the decoding of motor imagery by setting appropriate convolutional layers and adjusting network parameters. Results suggest that the performance of shallow convolutional neural networks (hereinafter referred to ConvNet) is comparative to classical methods utilizing FBCSP features and LDA classifiers. 
Another successful ANN architecture, EEGNet (\cite{lawhern2018eegnet}), has shown high robustness in various BCI paradigms includingerror-related negativity responses (ERN), P300 visual-evoked potentials, movement-related cortical potentials (MRCP), and sensory motor rhythms (SMR). Both ANN decoding methods work with the premise that the training data and the test data are sampled from the same distribution.
However, EEG signals are characterized by their non-stationary nature and low signal-to-noise ratio (\cite{sanei2013eeg}). Even when recorded from the same participant, the distribution of EEG features may exhibit discrepancies across different recording sessions (\cite{shenoy2006towards}).
In practical applications, cross-session EEG data collection is very common, participants are instructed to collect data of two (or more) sessions with time intervals. The data of several earlier sessions are used for training the model, and the data of the later sessions are used to evaluate the model performance. 

In the research field of machine learning, domain adaptation technologies are exploited to tackle the data non-stationary problem. Domain adaptation is a subcategory of transfer learning to solve the problem where source domain and target domain share the same feature space with different feature distribution. It helps transfer knowledge from source domain to target domain by learning domain invariants (\cite{zhang2020multimodel, zhang2020width}). The application of transfer learning techniques in MI-BCI research has been shown to enhance BCI performance in some studies. For example, \cite{he2019transfer}  proposed a pre-processing method for CSP filter, in which a spatial covariance matrix was used as a domain invariant to achieve trial-level EEG alignment in Euclidean space among different participants. \cite{zhao2020deep} proposed a GAN-based network  to improve the classification performance in the target participant by using data of other participants in the same dataset. \cite{zhang2021adaptive} proposed a variety of adaptive schemes by pre-training convolutional neural networks (CNN) models using EEG data from other participants and fine-tuning the model for one targeted participant. These studies utilized transfer learning techniques to enhance the classification performance of target participants by leveraging EEG data from other participants. However, considering the fact that a non-neglectable portion (15\%-30\%) (\cite{vidaurre2011machine}) of the population is BCI illiterate (\cite{allison2010could}), the data distribution of different participants and different sessions are quite different. Cross-participant domain adaptation methods on cross-session EEG data may introduce negative transfer (\cite{smith2001transfer}) and compromise the performance of MI classification. 

In this study, we approached the problem of cross-session variability in motor imagery data from the same participant by leveraging domain adaptation techniques to reduce the generalization error of artificial neural networks. A Siamese deep domain adaptation framework (SDDA) was proposed, with a universal configuration that can be easily attached to most existing neural networks. 
The manual domain invariants construction method was used as data preprocessing and integrated in SDDA as the first step to preliminarily reduce the distribution discrepancy between source domain and target domain. 
After the feature extraction layers, a maximum mean discrepancy (MMD) (\cite{long2015learning}) loss was integrated into the training loss function as regularization to further reduce the distribution difference in the Reproducing Kernel Hilbert Space (RKHS) (\cite{rosipal2001kernel}). 
A cosine-based center loss was introduced to reduce the adverse effects of noise and outliers on artificial neural networks. The SDDA framework was investigated using two commonly employed vanilla networks in EEG processing applications, EEGNet and ConvNet. Performance of the proposed SDDA was validated through experiments on public MI datasets from BCI competition IV (IIA and IIB) (\cite{tangermann2012review}). The proposed SDDA framework can significantly enhance the performance of vanilla networks, leading to superior classification accuracy both on IIA and IIB datasets.

The main contributions of this paper are:
\begin{enumerate}
    \item To the best of our knowledge, this research is the first deep domain adaptation work in the field of MI that does not require the assistance of other participants' data for training. This study explores whether and how deep domain adaptation can address the drift of EEG data over time in motor imagery.
    \item A transferable SDDA framework for MI classification is proposed, which consists of a preprocessing method for constructing domain-invariant features, an MMD loss for reducing domain-wise high-dimensional feature distribution discrepancies, and a cosine metric center loss for suppressing outliers and noises.
    \item We evaluate the classification performance of two high-impact vanilla networks in the SDDA framework on two real MI datasets. The experimental results demonstrate that the SDDA framework not only significantly improves the classification performance of the vanilla networks but also surpasses state-of-the-art domain adaptation methods. Furthermore, the experimental results highlight the significance of addressing the problem of cross-session variability in MI from the perspective of domain adaptation.
\end{enumerate}

The rest of this paper is organized in seven sections. In Section \ref{sec2}, we mainly review the relevant studies on deep learning and how domain adaptation technologies are used in MI EEG classiﬁcation. In Section \ref{sec3}, we explain the proposed SDDA framework in detail. The experiments and their results are presented and discussed in Section\ref{sec4}. In-depth analysis and discussion are presented in Section \ref{sec5}. Section \ref{sec6} discusses the benefits and limitations of the proposed SDDA framework. Finally, conclusions are summarized in Section \ref{sec7}.

\section{Related Works} \label{sec2}
Classical MI EEG classification process is generally composed of manual feature extraction and classification (\cite{baig2020filtering}). Common spatial pattern (CSP) and its variants, such as filter bank CSP (FBCSP) (\cite{ang2008filter}), Composite CSP (CCSP) (\cite{Kang2009}) and Stationary Subspace CSP (SSCSP) (\cite{Samek2013}) have played a vital role in feature extraction of motor imagery signals.
The foundational CSP algorithm is a spatial filter construction method tailored for two-class classification tasks, aiming to maximize inter-class variance through the development of a common spatial filter. FBCSP overcomes the single-frequency band limitation inherent in CSP, while CCSP and SSCSP integrate transfer learning techniques with CSP to enhance its feature extraction capabilities for MI signals. In the literature, linear discriminant analysis (LDA) and support vector machines (SVM) are the most prevalent classifiers employed in tandem with spatially filtered features to form a comprehensive MI-BCI decoding system.

Since ConvNet (\cite{schirrmeister2017deep})  proved that the end-to-end neural network can achieve the similar results as FBCSP in decoding MI, some researchers have focused MI decoding on the neural network (\cite{lawhern2018eegnet}, \cite{zhao2020deep}, \cite{craik2019deep}). Although researches of computer vision and natural language processing provide a solid foundation for MI decoding, the intrinsic properties of EEG are fundamentally different from those of images and texts. Research on ANN-based decoding methods for MI-EEG can be broadly classified into two distinct directions.

\paragraph{1) CNNs}\mbox{}

\emph{Finetuning CNN hyperparameters}: 
Research in the field of computer vision has demonstrated that convolution kernel size and the number of convolution layers play crucial roles in determining model performance. In the context of MI-BCI applications, EEGNet (\cite{lawhern2018eegnet}) and ConvNet (\cite{schirrmeister2017deep}) have been specifically designed with time-spatial convolutional layers and fine-tuned network hyperparameters to achieve optimal performance. The architectures of both ConvNet and EEGNet are relatively simple, eliminating the need for morphological conversion or data augmentation of EEG signals. This streamlined design contributes to the effectiveness and efficiency of these networks in decoding EEG signals for various applications.

\emph{Fusion prior knowledge with CNN}:
Some studies suggest that incorporating prior knowledge in the form of various features, such as time-space or time-frequency characteristics, from EEG signals might assist CNNs in capturing higher-level EEG representations more effectively. For example,  \cite{sakhavi2018learning} converted MI data into a temporal representation inspired by FBCSP algorithm, and designed specific neural network parameters for each participant for further feature extraction and classification. \cite{mammone2020deep} transformed MI data into a series of time-frequency maps using wavelet decomposition, and then fed the data into a CNN network to classify motor intention.

\emph{Complex ANN network architecture}:
Some studies attempt to extend the architecture of the basic convolutional layers to enhance the performance of the network. For example, \cite{sun2020eeg} proposed squeeze-and-excitation (SE) blocks to adaptively recalibrate the channel-wise features, enhancing the performance of CNN network in EEG classification.  \cite{amin2019deep} proposed a multi-scale CNN algorithm which could improve the accuracy of EEG MI classification by fusing the features extracted from different layers of CNN with convolution kernel.

\paragraph{2) GANs}\mbox{}

Generative adversarial networks (GANs) are able to learn joint data distributions, which has revolutionized the computer vision research field(\cite{ganin2016domain}, \cite{salimans2016improved}, \cite{arjovsky2017wasserstein}). In MI classification, \cite{zhao2020deep} proposed an end-to-end network model based on GAN, which successfully improved the classification performance by using data from other participants in the same dataset. \cite{fahimi2020generative} proposed a generative model based on deep convolutional generative adversarial network (DCGAN), which generated artificial EEG signals to increase the number of training data and subsequently improved the performance of MI-EEG classification. 

Deep domain adaptation is also a hot spot of computer vision research. Deep features in CNN are generally considered to be transitioning from general to specific in the last layers of the network (\cite{yosinski2014transferable}), and the classification layers which are tailored to specific tasks are hard to transfer from the source domain to the target domain directly. To keep features consistent between source domain and target domain, \cite{tzeng2014deep} introduced MMD loss as the domain confusion loss to learn a domain invariant representation. \cite{long2017deep} introduced a joint maximum mean discrepancy (JMMD) loss to align the joint distributions of multiple domain-specific layers across domains. \cite{rozantsev2018beyond} proposed a two-stream architecture and the weights of the streams were related by introducing the MMD loss. 

ASeveral studies have been reported in the literature concerning MI-EEG classification based on transfer learning techniques. For example, 
\cite{zhao2020deep} and \cite{zhang2021adaptive} both took the information from other participants into the consideration to improve the classification performance on a target participant.  However, the distributions of MI-EEG data can vary significantly across different participants (\cite{allison2010could}). In this reason,  \cite{zhao2020deep} and \cite{zhang2021adaptive} essentially solved the problem of multi-source domain adaptation(\cite{sun2015survey}). Furthermore, EEG data is often scarce, while neural network models typically require large amounts of data. Consequently, increasing the volume of training data can enhance the learning capabilities of these models. However, it remains unclear whether the improvements in model performance are attributable to the increased amount of data or the application of domain adaptation methods. In addition, due to the ethical concerns surrounding EEG data, leveraging data from other participants to assist in training may be limited in applications that are particularly sensitive to privacy issues.

In practical application scenarios, training data and test data from the same participant are always acquired with certain time intervals. Although there is evidence indirectly pointing out that data distributions of the same participant sampled with time intervals (cross-session) are not consistent (\cite{shenoy2006towards}), whether cross-session variability in MI can be treated as a domain adaptation is still unclear. In addition,  clear technical implementation for achieving domain adaptation in cross session MI EEG based BCI is also missing.

\section{Method} \label{sec3}
In this section, the mathematical notations and definitions used in this paper are firstly introduced. Then the domain adaptation theory and mathematical models are applied to the cross-session MI-EEG classification. The SDDA framework is introduced from the perspective of reducing the domain distribution bias and reduce generalization error of the source domain classifiers. Finally, from the perspective of backpropagation, the rationale analysis of the loss function is presented.

\subsection{Notations and Definitions}
In this paper, the data recorded from training sessions are noted as the source domain, 
$ \mathcal{D}^{s}=\left\{\mathbf{X}^{s}, \mathbf{Y}^{s}\right\}=\left\{\left(\mathbf{x}_{i}^{s}, y_{i}^{s}\right)\right\}_{i=1}^{n_{s}} $  
and the unlabeled test sessions are regarded as target domain,  
$ \mathcal{D}^{t}=\left\{\mathbf{X}^{t}\right\}=\left\{\mathbf{x}_{i}^{t}\right\}_{i=1}^{n_{t}} $ , 
where  $\mathbf{x}_{i}^{s} \in \mathbb{R}^{E \times T}$ denotes the $ith$ source trial data in total  $n_s$ source data sampled from $E$ electrode channels and $T$ time samples during the available MI duration in a trial and  $ y_{i}^{s} \in \mathbb{R}^{c} $  denotes the corresponding class of  $\mathbf{x}_{i}^{s}$ in total $C$ classes.  Parallel notations are used for unlabeled target data $\mathbf{x}_{i}^{t}$. 

Since we hypothesis cross-session MI classification as a domain adaptation problem, the mathematical representation of generalization error is firstly derived under domain adaptation theory.
\newtheorem{thm}{Theorem}
\begin{thm}  
For a hypothesis h,
\begin{align*}
    &\epsilon^{t}(h) \leq  \epsilon^{s}(h)+d_{1}\left(\mathcal{D}^{s}, \mathcal{D}^{t}\right)+ \\
    &\min \left\{\mathrm{E}_{D^{s}}\left[\left|f^{s}(\mathbf{x})-f^{t}(\mathbf{x})\right|\right], \mathrm{E}_{D^{t}}\left[\left|f^{s}(\mathbf{x})-f^{t}(\mathbf{x})\right|\right]\right\} 
\end{align*}
where superscript ‘s’ and ‘t’ indicate the items from the source domain and target domain, $ \epsilon^{s}(h) $ is the source error, that a model training in the labeled source domain might seek to minimize. $d_{1}\left(\mathcal{D}^{s}, \mathcal{D}^{t}\right)$  is the $L_1$  divergence between $\mathcal{D}^{s}$ and  $\mathcal{D}^{t}$. $f(\mathbf{x})$ is the labeling function of input $\mathbf{x}$ . (\cite{ben2010theory})
\label{theorem1}
\end{thm}

It can be seen from Theorem \ref{theorem1} that the target error depends on the source error, $L_1$  divergence between the source domain and target domain, as well as the performance difference of labeling functions between the source domain and target domain.  However,  Theorem \ref{theorem1} is cannot be directly used in practical scenarios, as $d_{1}\left(\mathcal{D}^{s}, \mathcal{D}^{t}\right)$ is too strict and cannot be accurately estimated from finite samples of domain distributions (\cite{Batu2000}). Due to the non-stationarity and low signal-to-noise ratio characteristics of EEG data, accurately measuring its distribution poses a great challenge. Fortunately, domain adaptation can be performed in feature representations under the framework of Vapnik-Chervonenkis (VC) dimension (\cite{Vapnik1994}), even with finite samples.

\begin{thm}
    Let $\mathcal{R}$ be a fixed representation function from $\mathcal{X}$ to $\mathcal{Z}$ and $\mathcal{H}$ be a hypothesis space of VC-dimension $d$. If a random labeled sample of size $m$ is generated by applying $R$ to a $\mathcal{D}^{s}$ $i.i.d$ sample labeled according to $f$ , then with probability at least $1-\delta$ ,  for every  $h \in \mathcal{H}$:
\begin{align*}
    \epsilon^{t}(h) \leq \hat{\epsilon}^{s}(h)&+ \sqrt{\frac{4}{m}\left(d \log \frac{2 e m}{d}+\log\frac{4}{\delta}\right)} \\ & +d_{\mathcal{H}}\left(\tilde{\mathcal{D}}^{s},\tilde{\mathcal{D}}^{t}\right)
    +\lambda
\end{align*}
where $e$ is the base of the natural logarithm, $\lambda$ is a parameter which indicates the complexity of the labeling function under the $\mathcal{A}$-distance.  In other words, a small $\lambda$ indicates our induced labeling function performs well on both domains.(\cite{Ben-David2007})
\label{theorem2}
\end{thm}

Noticed that, in the case of arbitrarily distribution of finite data, it is difficult to accurately measure $d_{\mathcal{H}}$. \cite{Ben-David2007} used  $\mathcal{A}$-distance  to measure the difference between the $\tilde{\mathcal{D}}^{s}$ and $\tilde{\mathcal{D}}^{t}$ , but it is still a NP-hard problem even to approximate the error of the optimal hyperplane classifier for arbitrary distributions (\cite{kifer2004detecting}). For non-stationary EEG signals with low signal-to-noise ratio, it is even more difficult to accurately express $d_{\mathcal{H}}$. However, combining Theorem \ref{theorem1} and Theorem \ref{theorem2}, domain adaptation can be achieved by minimizing either the original data distributions or the representation discrepancy. With a pre-defined dataset, to reduce the error in target domain, the error in source domain and the difference between the source domain and the target domain should be minimized jointly. 

In the context of MI decoding, most existing studies assume that the data from the source and target domains have the same distribution, overlooking the data distribution differences caused by cross-session variability. Therefore,
these methods only focused on minimizing  $ \epsilon^{s}(h) $  or $ \tilde{\epsilon}^{s}(h) $. 
To this end, we fully considered the factors that affect $ \epsilon^{t}(h)$, and further investigated the impact of the distribution differences in both the raw data and high-dimensional feature representations of the source and target domains on the cross-session variability problem, while attempting to reduce $ \tilde{\epsilon}^{s}(h)$.

\begin{figure*} 
\centering 
\includegraphics[width=0.85\textwidth]{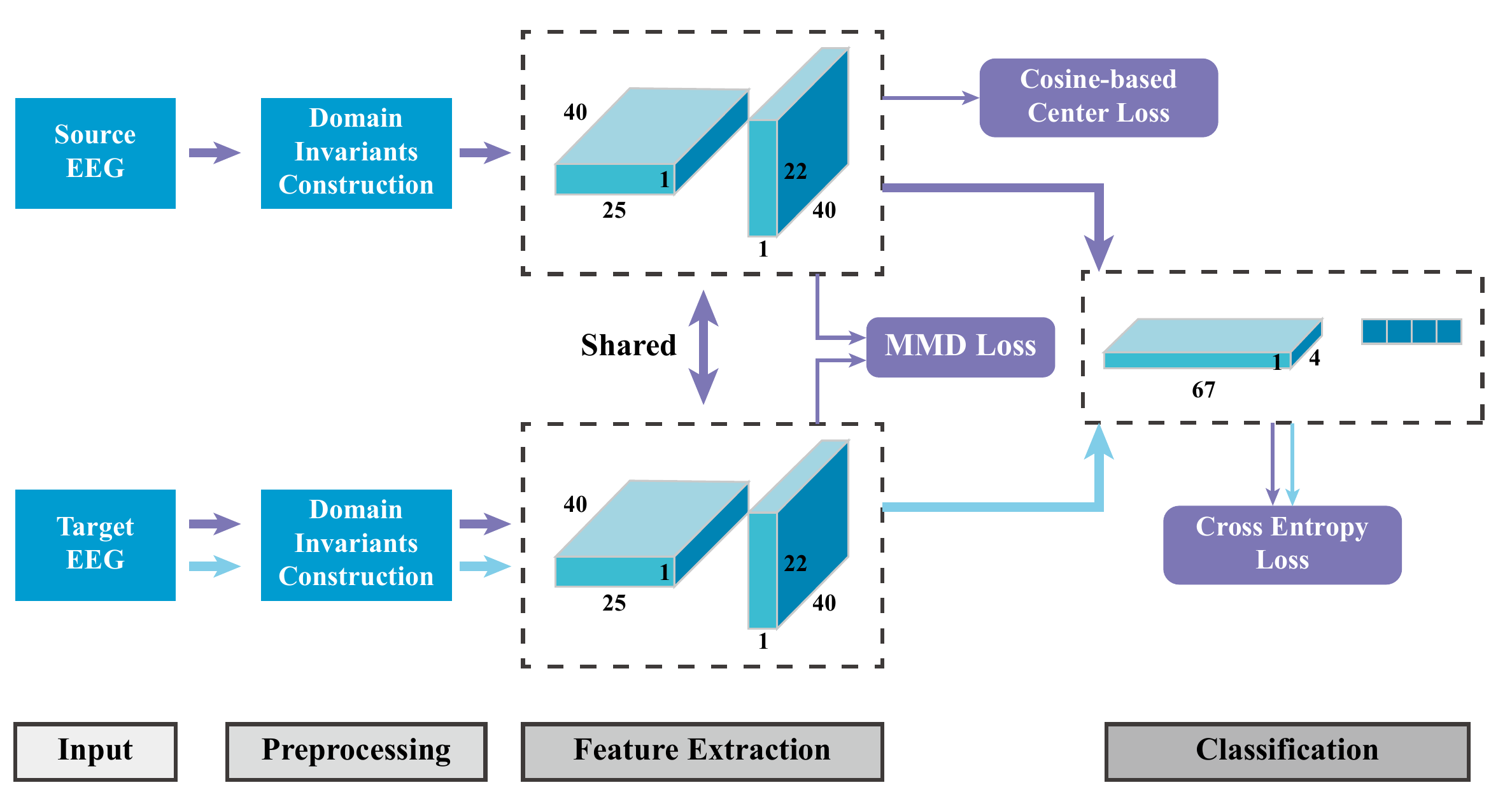} 
\caption{Framework of the proposed Siamese deep domain adaptation (SDDA) network. The purple arrow indicates the training process and the blue arrow indicates the testing process. SDDA is an universal framework, which is compatible to any neural network with feature extraction layers and classification layers. Feature extraction layers and classification layers of ConvNet (\cite{schirrmeister2017deep}) can be embedded into the corresponding dotted line box in the figure to get ConvNet-based SDDA framework, named DA-ConvNet below. "Preprocessing" in the figure includes channel normalization  and Euclidean alignment in \ref{preprocessing}. MMD loss and cosine-based center loss are implemented by the method in \ref{sec:mmd} and \ref{sec:cosine}, respectively. The cross entropy loss is used for classification in equation \eqref{cross entropy}.}
\label{Fig.framework} 
\end{figure*}

The proposed SDDA framework is presented in Fig. \ref{Fig.framework}, which consists of two main branches with shared parameters, one for each domain. 
Let $\mathbf{\Theta}$ denote the parameters in Siamese network, $\mathbf{\Theta}_1$ denote the shared parameters of feature extraction layers. $\mathbf{H}^{s} \in \mathbb{R}^{b\times L} $  and $\mathbf{H}^{t} \in \mathbb{R}^{b\times L} $  denote embedding features outputting from the feature extraction layers from the source domain and target domain in a mini-batch. Here, $b$ indicates the mini-batch size, and $L$ is the length of the embedding features, which also represents the size of inputs to the classification layers.

The overall framework can be trained by minimizing the following loss function.
\begin{align}
\label{eq1}
\mathcal{L}\left(\mathbf{\Theta} \mid \mathbf{X}^{s}, \mathbf{Y}^{s},  
\mathbf{X}^{t}\right)&=\mathcal{L}_{\mathrm{s}}+\lambda_{1} \mathcal{L}_{c}+\lambda_{2} \mathcal{L}_{d} \\  
\mathcal{L}_{s}&=\mathcal{J}_{s}\left(\mathbf{\Theta} \mid \mathbf{X}^{s}, \mathbf{Y}^{s}\right) \\                   
\mathcal{L}_{c}&=\mathcal{J}_{c}\left(\mathbf{\Theta}_{1} \mid \mathbf{X}^{s}, \mathbf{Y}^{s}\right) \\             
\mathcal{L}_{d}&=\mathcal{J}_{d}\left(\mathbf{\Theta}_{1} \mid \mathbf{X}^{s}, \mathbf{X}^{t}\right)                 
\end{align}
where $\mathcal{L}_{s}$, $\mathcal{L}_{c}$ and $\mathcal{L}_{d}$ represent the cross entropy loss, cosine-based center loss and MMD loss,  respectively. $\lambda_1$ and $\lambda_2$ are trade-off parameters to balance the contribution of $\mathcal{L}_{c}$ and $\mathcal{L}_{d}$, respectively.

\subsection{Domain discrepancy minimization}
Given the low signal-to-noise ratio characteristic of EEG data, preprocessing of source and target domain data is essential before inputting it into an artificial neural network for feature extraction and classification. To address this, we propose an EEG data preprocessing method tailored for artificial neural networks, aiming to reduce domain distribution differences. The proposed method initially normalizes the data of trials in different domains using [-1, 1] normalization. Subsequently, the average covariance matrix of the source and target domains is adjusted through Euclidean alignment (\cite{he2019transfer}). 

\subsubsection{Preprocessing} \label{preprocessing}

\paragraph{1) Trial normalization} \label{sec:channel} \mbox{} \\
Let $\mathbf{x}_{i}$ denote the $i$-th EEG trial, and each trial's amplitude is normalized to the range [-1, 1] using the following equation:
\begin{equation} 
\mathbf{x}_{i}=\frac{\mathbf{x}_{i}}{\max \left(\left|\mathbf{x}_{i}\right|\right)}    
\end{equation}
where $\mathbf{x}_{i}$ denotes the $i$th trial of $\mathbf{X}$ for both source domain and target domain,  $\left|\cdot\right|$ denotes taking the absolute value of the matrix.

\paragraph{2) Euclidean Alignment} \label{sec:euclidean} \mbox{} \\
Following trial-wise normalization, Euclidean alignment (\cite{he2019transfer}) was applied by calculating the mean covariance matrix for each domain. Originally employed in a cross-participant motor imagery classification scenario to obtain more robust CSP filters, the Euclidean alignment method has been adapted in this work as an integral component of the SDDA framework for MI-EEG cross-session classification. Assuming a domain consists of $n$ trials, the arithmetic means of all covariance matrices, calculated using EEG data from the same domain, are determined as follows:
\begin{equation}
    \bar{R}=\frac{1}{n} \sum_{i=1}^{n} \mathbf{x}_{i} \mathbf{x}_{i}^{T}
\end{equation}
Therefore, Euclidean alignment can be performed as: 
\begin{equation}
    \tilde{\mathbf{x}}_{i}=\bar{R}^{-\frac{1}{2}} \mathbf{x}_{i}
\end{equation}

It should be noted that the domain adaptive EEG data pre-processing method proposed in this study is not limited by decoding methods and can be integrated into any manual feature extraction or artificial neural network methods. 

\subsubsection{Reproducing Kernel Hilbert Space Alignment} \label{sec:mmd}
Since some studies (\cite{gretton2012kernel}, \cite{long2017deep}, \cite{tzeng2014deep}) had demonstrated the effectiveness of MMD in adjusting domain distributions, MMD loss was also used as a regularization term to reduce the high-dimensional feature discrepancies between source and target domains in SDDA  framework. 

 Assuming the mapping function in MMD is a unit ball in Reproducing Kernel Hilbert space (RKHS) $\mathcal{H}_{k}$ (\cite{rosipal2001kernel}), the MMD loss is defined as: 
\begin{equation}
    d_{k}^{2}(p, q) \triangleq \left\|\mathbf{E}_{p}\left[\phi\left(\mathbf{X}^{s}\right)\right]-\mathbf{E}_{q}\left[\phi\left(\mathbf{X}^{t}\right)\right]\right\|_{\mathcal{H}_{k}}^{2}  
\end{equation}
where $\mathbf{E}_{p}$  and $\mathbf{E}_{q}$  represent the mean embedding of probability distribution $p$ and $q$. $\phi\left(\mathbf{X}\right)$  represents the feature mapping of a domain.  The most important property for MMD loss is that $p=q$ if $d_{k}^{2}(p, q)=0$, which is exactly the property we expect in Theorem \ref{theorem2}.

In this paper, Gaussian kernel was used to map the features into a unit ball in a universal RKHS. 

\subsection{Learning Discriminative features} \label{sec:cosine}
Based on previous literature (\cite{wen2016discriminative}), it has been shown that the center loss can effectively reduce the Euclidean distance between learned features and their corresponding class centers, thereby encouraging samples from the same class to cluster closely in the embedding feature space.
However, Euclidean distance methods are generally considered vulnerable to noises and outlier samples (\cite{wang2018cosface}), which are very common in EEG due to its non-stationary and low signal-to-noise ratio character. To solve this issue, cosine-based center loss was proposed, which transformed the Euclidean distance measure into cosine based distance. The cosine-based center loss is calculated as:
\begin{align}\begin{aligned}  
\label{center}
\mathcal{L}_{c}&=\frac{1}{2 b} \sum_{i=1}^{b}\left\|\bar{\mathbf{h}}_{i}^{s}-\bar{\mathbf{c}}_{y_j}\right\|_{2}^{2} ,\\
\text{where,} & \qquad \bar{\mathbf{h}}_{i}^{s}=\frac{\mathbf{h}_{i}^{s}}{\left\|\mathbf{h}_{i}^{s}\right\|_{2}} \\
\qquad &\qquad \bar{\mathbf{c}}_{y_{j}}=\frac{\mathbf{c}_{y_{j}}}{\left\|\mathbf{c}_{y_{j}}\right\|_{2}}
\end{aligned}\end{align}
In equation \ref{center}, $\mathbf{h}_{i}^{s}$  denotes the embedding features of the $ith$ sample in the source domain, $\mathbf{c}_{y_{j}}$   denotes the center of the $jth$ label of the embedding features in the source domain, $b$ denotes the mini-batch size, and $\left\| \cdot \right\|_{2}$  represents the Euclidean norm.

By adding the constraints, equation (\ref{center}) can be expressed as follows: 
\begin{align}\begin{aligned}  
\label{cosine}
\mathcal{L}_{c}=&\frac{1}{2 b} \sum_{i=1}^{b}\left\|\bar{\mathbf{h}}_{i}^{s}-\bar{\mathbf{c}}_{y_{j}}\right\|_{2}^{2} \\
=& \frac{1}{2 b} \sum_{i=1}^{b}\left\|\bar{\mathbf{h}}_{i}^{s}\right\|_{2}^{2}+\left\|\bar{\mathbf{c}}_{y_{j}}\right\|_{2}^{2}-2\left<\bar{\mathbf{h}}_{i}^{s}, \bar{\mathbf{c}}_{y_{j}}\right>\\
=& 1-\frac{1}{b} \sum_{i=1}^{b} \cos \left(\mathbf{h}_{i}^{s}, \mathbf{c}_{y_{j}}\right)
\end{aligned}\end{align}
According to Equation (\ref{cosine}), the distance between the embedding features and their corresponding center is measured by a cosine function. $\mathcal{L}_{c}$  is inherently bounded between [0, 2]. That means if the embedding features are close to the corresponding center, the cosine value will be close to 1, making $\mathcal{L}_{c}$  close to 0. However, if the embedding features are far from the corresponding center,  $\mathcal{L}_{c}$ demonstrates a limited response, which restricts the maximum impact of the outlier data.
\subsection{Training}
Cross entropy loss (\cite{zhang2018generalized}) was used for classification:

\begin{equation} 
    \mathcal{L}_{s}=-\sum_{i=1}^{b} \log \frac{e^{W_{y_{i}}^{T} \mathbf{h}_{i}+b_{y_{i}}}}{\sum_{j=1}^{n} e^{W_{j}^{T} \mathbf{h}_{i}+b_{j}}}
    \label{cross entropy}
\end{equation}
where $n$ represents the number of motor imagery classes. 

Then the network can be trained by minimizing the Equation (\ref{eq1}). $\mathcal{L}_{s}$, $\mathcal{L}_{c}$  and $\mathcal{L}_{d}$   in equation (\ref{eq1}) are all differentiable $w.r.t.$ the inputs of the network. Therefore, the network parameters can be updated by the standard backpropagation. 

\begin{equation} 
    \mathbf{\Theta}^{l+1}=\mathbf{\Theta}^{l}-\eta \frac{\partial\left(\mathcal{L}_{s}+\lambda_{1} \mathcal{L}_{c}+\lambda_{2} \mathcal{L}_{d}\right)}{\partial \mathbf{x}_{\mathbf{i}}}
\end{equation}
where $\eta$ is the learning rate to update the network parameters and $l$ is the number of epochs during training iterations. All three losses above are implemented via the mini-batch strategy.The global class center has its own parameters which update simultaneously with the network parameters in each training step.

\begin{equation} 
    \mathbf{c}_{j}^{l+1}=\mathbf{c}_{j}^{l}-\gamma \cdot \Delta \mathbf{c}_{j}^{l} \quad j=1,2, \cdots, c
\end{equation}
where $\gamma$ is the learning rate to update the parameters of center loss.

\section{Experiments} \label{sec4}
In the following section, we evaluate the performance of the proposed framework through experiments conducted on two publicly available MI EEG datasets: dataset IIA (four-class MI) and dataset IIB (binary class MI with sparse channels) from BCI Competition IV (Tangermann et al., 2012). We selected two popular CNN architectures for EEG applications, namely EEGNet (\cite{lawhern2018eegnet})  and ConvNet (\cite{schirrmeister2017deep}), to verify the performance of the proposed framework. Furthermore, we compare the performance of the proposed domain adaptation framework (SDDA) with that of the tested baseline CNN models and several representative methods to provide a comprehensive assessment of its efficacy.

\subsection{Datasets}
\subsubsection{BCI Competition IV Dataset IIA}
In IIA dataset\footnote{http://www.bbci.de/competition/iv/\#dataset2a}, EEG was collected from a twenty-two channel EEG with 10-20 configuration at a sampling rate of 250 Hz from nine healthy participants (ID A01-A09) in two different sessions with multiple days’ interval. Each participant participated in four motor imagery tasks, including imagining the movement of left hand, right hand, both feet and tongue. Each session contained 288 trials of EEG data and all data collected in the first acquisition session were used for training (source domain) and those acquired in the second session were used for testing (target domain).
According to ConvNet (\cite{schirrmeister2017deep}), temporal segmentation of [1.5, 6] seconds after each MI cue was extracted as one trial of EEG data.  

\subsubsection{BCI Competition IV Dataset IIB}
In IIB dataset\footnote{http://www.bbci.de/competition/iv/\#dataset2b}, EEG was collected with a three EEG electrode channels sampled at 250 Hz from nine healthy participants (ID B01-B09) in five separate acquisition sessions with multiple days’ interval. Each participant participated in two motor imagery tasks, including imagining the movement of left hand and right hand. The first two sessions contained 120 trials per session without feedback and the last three sessions contained 160 trials per session with a smiley face on the screen as feedback. 
As described in \cite{zhao2020deep}, all data in the first three sessions were used for training (source domain) and the last two sessions were used for test (target domain). Temporal segmentation of [2, 5] seconds after MI cue was extracted as one trial of EEG data in our experiment.

\subsection{Filtering and Network hyperparameters}

In this paper, a 200-order Blackman window bandpass FIR filter with a [4, 38] Hz range was employed to filter the EEG data. Given that EEGNet and ConvNet have been extensively researched in this area, both models were selected as the baseline networks, with their classification performance on datasets IIA and IIB serving as the benchmarks for this study. The detailed network architecture and parameters of ConvNet and EEGNet are presented in Table \ref{table:convnet} and Table \ref{table:eegenet}, respectively. Subsequently, EEGNet and ConvNet were trained in conjunction with the proposed SDDA framework, and their classification performance was reported to illustrate the rationale and feasibility of domain adaptation in cross-session MI EEG classification. To ensure a fair comparison, preprocessing, network hyperparameters, and training strategies were maintained consistent with the original baseline networks. The optimal performance of different networks was recorded to minimize the influence of random factors. Moreover, we conducted all experiments five times and reported the mean accuracy to further reduce randomness. Our baseline classification results for vanilla ConvNet on the two datasets were consistent with those reported by \cite{zhao2020deep}. However, our results for vanilla EEGNet varied significantly, which may be attributed to changes in the EEGNet architecture across different publication versions. In this paper, we adopted the network architecture of EEGNet as published by its authors on Github \footnote{\url{https://github.com/vlawhern/arl-eegmodels}}.

The detailed network configuration and hyperparameters are shown in Table \ref{table:convnet} and Table \ref{table:eegenet}. To the left of the slashes in the tables are the amount of parameters of the network in the IIA dataset, and to the right are the amount in the IIB dataset. 

\begin{table}[H!]
\caption{Model Parameters of  ConvNet }
\centering
\label{table:convnet}
\resizebox{\linewidth}{!}{
\begin{tabular}{|c|c|c|c|}
\hline
Modules                            & Layers                           & Parameters                                                                           & Num of param       \\ \hline
\multirow{8}{*}{\begin{tabular}[c]{@{}c@{}}Feature\\    \\ Extractor\end{tabular}} 
                                   & Temporal Conv                    & (1, 25), 40                                                                    & 1040               \\ \cline{2-4} 
                                   & Spatial Conv                     & (E, 1), 40                                                                     & 35200/4800         \\ \cline{2-4} 
                                   & BatchNorm                        & -                                                                              & 80                 \\ \cline{2-4} 
                                   & Square                & -                                                                              & -                  \\ \cline{2-4} 
                                   & Average Pooling                 & (1, 75),15 
                                                                      & -            \\ \cline{2-4} 
                                   & Logarithm             & -                                                                              & -                  \\ \cline{2-4} 
                                   & Dropout                          & p=0.5                                                                          & -                  \\ \hline
\multirow{1}{*}{Classifier}        & Conv2d                           & (1, N), C                                                                      & 11044/4162         \\  \hline
Total   & -
                          & -                                                                              & 47364/10082        \\ \hline
\end{tabular}}
\end{table}

\begin{table}[H!]
\caption{Model Parameters of EEGNet}
\centering
\resizebox{\linewidth}{!}{
\label{table:eegenet}
\begin{tabular}{|c|c|c|c|}
\hline
Modules                                                                             & Layers          &Parameters    &Num of Param\\ \hline
\multirow{12}{*}{\begin{tabular}[c]{@{}c@{}}Feature\\    \\ Extractor\end{tabular}} & Temporal Conv   & (1,   64), 8  & 512          \\ \cline{2-4} 
                                                                                    & BatchNorm       & -             & 16           \\ \cline{2-4} 
                                                                                    & Depthwise Conv  & (E,   1), 8   & 176/24       \\ \cline{2-4} 
                                                                                    & BatchNorm       & -             & 16/          \\ \cline{2-4} 
                                                                                    & ELU Action      & -             & -            \\ \cline{2-4} 
                                                                                    & Average Pooling & (1,   4)      & -            \\ \cline{2-4} 
                                                                                    & Dropout         & p=0.5         & -            \\ \cline{2-4} 
                                                                                    & Separable Conv  & (1,   16), 16 & 272/128      \\ \cline{2-4} 
                                                                                    & BatchNorm       & -             & 32           \\ \cline{2-4} 
                                                                                    & ELU Action      & -             & -            \\ \cline{2-4} 
                                                                                    & Average Pooling & (1,   8)      & -            \\ \cline{2-4} 
                                                                                    & Dropout         & p=0.5         & -            \\ \hline
\multirow{1}{*}{Classifier}                                                         & Fully Connected & C             & 1988/738     \\ 
 \hline
Total   & -        & -             & 3012/1610    \\ \hline
\end{tabular}}
\end{table}

\subsection{Experiment Settings}
All the experiments were conducted under the Pytorch framework on a PC with an Intel(R) Xeon(R) Gold 5117 CPU @ 2.00 GHz and Nvidia Tesla V100 GPU. For the above two datasets, all EEG channels were used for classification, and the three electrooculography channels were not included in the analysis. 

AdamW (\cite{loshchilov2018fixing}) was used as the optimizer to train the proposed SDDA framework. The learning rates were set to 0.0001 for ConvNet and 0.001 for EEGNet. The networks were trained with mini-batches with size of 16. The trade-off parameters were grid-searched with $\lambda_1$ from \{0, 0.2, 1, 2, 10, 15\} and $\lambda_2$ from \{0, 0.02, 0.05, 0.1, 0.2, 0.5\} for each participant. The training strategies were the same with those in ConvNet (\cite{schirrmeister2017deep}). In the first training stage, the data in the source domain were randomly sampled 20\% as validation data to monitor the training process to avoid overfitting. And in the second training stage, all the data in the source domain were used for training.

In addition to the vanilla EEGNet and vanilla ConvNet, the SDDA framework were also compared with a variety of successful methods in the literature. FBCSP (\cite{ang2008filter}), CCSP (\cite{Kang2009}) and SSCSP (\cite{Samek2013}) are considered as traditional feature extraction methods based on spatial patterns. DRDA (\cite{zhao2020deep}) is a GAN-based ANN method. The Kappa value ($\kappa$) (\cite{cohen1960coefficient}) was used as accuracy measure in this study, which is calculated as:
\begin{equation}
    \kappa=\frac{acc-p_{0}}{1-p_{0}}
\end{equation}
where $acc$ is the standard accuracy, and $p_0$ is the random level accuracy.
\begingroup
\begin{table*}
\caption{Classification Accuracy (\%) of Different Algorithms on Dataset IIA of BCI Competition IV}
\centering
\label{table3}
\begin{threeparttable}  
\begin{tabular}{|c|ccccccccc|c|}
\hline
                         & \multicolumn{9}{c|}{Participant   ID}                                                                                                                                                                                                                                                                                                                                                                                                         &                                                                                   \\ \cline{2-10}
\multirow{-2}{*}{Method} & \multicolumn{1}{c|}{A01}                           & \multicolumn{1}{c|}{A02}                           & \multicolumn{1}{c|}{A03}                           & \multicolumn{1}{c|}{A04}                           & \multicolumn{1}{c|}{A05}                           & \multicolumn{1}{c|}{A06}                           & \multicolumn{1}{c|}{A07}                           & \multicolumn{1}{c|}{A08}                           & A09   & \multirow{-2}{*}{\begin{tabular}[c]{@{}c@{}}Average   acc\\ (kappa)\end{tabular}} \\ \hline
FBCSP                    & \multicolumn{1}{c|}{76.00}                         & \multicolumn{1}{c|}{56.50}                         & \multicolumn{1}{c|}{81.25}                         & \multicolumn{1}{c|}{61.00}                         & \multicolumn{1}{c|}{55.00}                         & \multicolumn{1}{c|}{45.25}                         & \multicolumn{1}{c|}{82.75}                         & \multicolumn{1}{c|}{81.25}                         & 70.75 & 67.75(0.570)                                                                      \\ \hline
\#CCSP                   & \multicolumn{1}{c|}{84.72}                         & \multicolumn{1}{c|}{52.78}                         & \multicolumn{1}{c|}{80.90}                         & \multicolumn{1}{c|}{59.38}                         & \multicolumn{1}{c|}{54.51}                         & \multicolumn{1}{c|}{49.31}                         & \multicolumn{1}{c|}{88.54}                         & \multicolumn{1}{c|}{71.88}                         & 56.60 & 66.51(0.553)                                                                      \\ \hline
\#SSCSP                  & \multicolumn{1}{c|}{76.74}                         & \multicolumn{1}{c|}{58.68}                         & \multicolumn{1}{c|}{81.25}                         & \multicolumn{1}{c|}{57.64}                         & \multicolumn{1}{c|}{38.54}                         & \multicolumn{1}{c|}{48.26}                         & \multicolumn{1}{c|}{76.39}                         & \multicolumn{1}{c|}{79.17}                         & 78.82 & 66.17(0.548)                                                                      \\ \hline
\rowcolor[HTML]{9B9B9B} 
VA-EEGNet                & \multicolumn{1}{c|}{\cellcolor[HTML]{9B9B9B}75.34} & \multicolumn{1}{c|}{\cellcolor[HTML]{9B9B9B}51.04} & \multicolumn{1}{c|}{\cellcolor[HTML]{9B9B9B}88.54} & \multicolumn{1}{c|}{\cellcolor[HTML]{9B9B9B}57.29} & \multicolumn{1}{c|}{\cellcolor[HTML]{9B9B9B}46.52} & \multicolumn{1}{c|}{\cellcolor[HTML]{9B9B9B}50.34} & \multicolumn{1}{c|}{\cellcolor[HTML]{9B9B9B}83.68} & \multicolumn{1}{c|}{\cellcolor[HTML]{9B9B9B}80.55} & 87.15 & 68.94(0.585)                                                                      \\ \hline
\rowcolor[HTML]{D9D9D9} 
VA-ConvNet               & \multicolumn{1}{c|}{\cellcolor[HTML]{D9D9D9}82.29} & \multicolumn{1}{c|}{\cellcolor[HTML]{D9D9D9}57.63} & \multicolumn{1}{c|}{\cellcolor[HTML]{D9D9D9}92.70} & \multicolumn{1}{c|}{\cellcolor[HTML]{D9D9D9}\textbf{81.94}} & \multicolumn{1}{c|}{\cellcolor[HTML]{D9D9D9}55.2}  & \multicolumn{1}{c|}{\cellcolor[HTML]{D9D9D9}44.44} & \multicolumn{1}{c|}{\cellcolor[HTML]{D9D9D9}89.93} & \multicolumn{1}{c|}{\cellcolor[HTML]{D9D9D9}83.33} & 82.29 & 74.41(0.633)                                                                      \\ \hline
\rowcolor[HTML]{D9D9D9} 
\#DRDA                   & \multicolumn{1}{c|}{\cellcolor[HTML]{D9D9D9}83.19} & \multicolumn{1}{c|}{\cellcolor[HTML]{D9D9D9}55.14} & \multicolumn{1}{c|}{\cellcolor[HTML]{D9D9D9}87.43} & \multicolumn{1}{c|}{\cellcolor[HTML]{D9D9D9}75.28} & \multicolumn{1}{c|}{\cellcolor[HTML]{D9D9D9}\textbf{62.29}} & \multicolumn{1}{c|}{\cellcolor[HTML]{D9D9D9}57.15} & \multicolumn{1}{c|}{\cellcolor[HTML]{D9D9D9}86.18} & \multicolumn{1}{c|}{\cellcolor[HTML]{D9D9D9}83.61} & 82.00 & 74.75(0.663)                                                                      \\ \hline
\rowcolor[HTML]{D9D9D9} 
*DA-ConvNet(ours)              & \multicolumn{1}{c|}{\cellcolor[HTML]{D9D9D9}\textbf{90.62}} & \multicolumn{1}{c|}{\cellcolor[HTML]{D9D9D9}\textbf{62.84}} & \multicolumn{1}{c|}{\cellcolor[HTML]{D9D9D9}\textbf{93.40}} & \multicolumn{1}{c|}{\cellcolor[HTML]{D9D9D9}\textbf{84.02}} & \multicolumn{1}{c|}{\cellcolor[HTML]{D9D9D9}\textbf{68.05}} & \multicolumn{1}{c|}{\cellcolor[HTML]{D9D9D9}\textbf{61.80}} & \multicolumn{1}{c|}{\cellcolor[HTML]{D9D9D9}\textbf{97.20}} & \multicolumn{1}{c|}{\cellcolor[HTML]{D9D9D9}\textbf{90.97}} & \textbf{89.23} & \textbf{82.01(0.760)}                                                                      \\ \hline
\rowcolor[HTML]{9B9B9B} 
*DA-EEGNet(ours)                & \multicolumn{1}{c|}{\cellcolor[HTML]{9B9B9B}\textbf{88.54}} & \multicolumn{1}{c|}{\cellcolor[HTML]{9B9B9B}\textbf{69.09}} & \multicolumn{1}{c|}{\cellcolor[HTML]{9B9B9B}\textbf{97.22}} & \multicolumn{1}{c|}{\cellcolor[HTML]{9B9B9B}72.20} & \multicolumn{1}{c|}{\cellcolor[HTML]{9B9B9B}57.98} & \multicolumn{1}{c|}{\cellcolor[HTML]{9B9B9B}\textbf{59.72}} & \multicolumn{1}{c|}{\cellcolor[HTML]{9B9B9B}\textbf{93.06}} & \multicolumn{1}{c|}{\cellcolor[HTML]{9B9B9B}\textbf{86.80}} & \textbf{90.27} & \textbf{79.43(0.725)}                                                                      \\ \hline
\end{tabular}
 \begin{tablenotes}
        \footnotesize
        \item[] White, light grey and dark shade indicate CSP-based methods, ConvNet-based method and EEGNet-based method, respectively. Methods with ”\#” adjust the data distribution among participants. Methods with “*” adjust the data distribution within the same participants
      \end{tablenotes}
    \end{threeparttable}
\end{table*}

\begin{table*}
\caption[Caption for LOF]{Classification Accuracy (\%) of Different Algorithms on Dataset IIB of BCI Competition IV}
\centering
\label{table4}
\begin{threeparttable}  
\begin{tabular}{|c|ccccccccc|c|}
\hline
                                    & \multicolumn{9}{c|}{Participant   ID}                                                                                                                                                                                                                                                                                                                                                                                                                                                                                          &                                                                                   \\ \cline{2-10}
\multirow{-2}{*}{Method}            & \multicolumn{1}{c|}{B01}                                    & \multicolumn{1}{c|}{B02}                                    & \multicolumn{1}{c|}{B03}                                    & \multicolumn{1}{c|}{B04}                                    & \multicolumn{1}{c|}{B05}                                    & \multicolumn{1}{c|}{B06}                                    & \multicolumn{1}{c|}{B07}                                    & \multicolumn{1}{c|}{B08}                                    & B09            & \multirow{-2}{*}{\begin{tabular}[c]{@{}c@{}}Average   acc\\ (kappa)\end{tabular}} \\ \hline
FBCSP                               & \multicolumn{1}{c|}{70.00}                                  & \multicolumn{1}{c|}{60.36}                                  & \multicolumn{1}{c|}{60.94}                                  & \multicolumn{1}{c|}{97.50}                                  & \multicolumn{1}{c|}{93.12}                                  & \multicolumn{1}{c|}{80.63}                                  & \multicolumn{1}{c|}{78.13}                                  & \multicolumn{1}{c|}{92.50}                                  & 87.88          & 80.00(0.600)                                                                      \\ \hline
\#CCSP                              & \multicolumn{1}{c|}{63.75}                                  & \multicolumn{1}{c|}{56.79}                                  & \multicolumn{1}{c|}{50.00}                                  & \multicolumn{1}{c|}{93.44}                                  & \multicolumn{1}{c|}{65.63}                                  & \multicolumn{1}{c|}{81.25}                                  & \multicolumn{1}{c|}{72.81}                                  & \multicolumn{1}{c|}{87.81}                                  & 82.81          & 72.70(0.454)                                                                      \\ \hline
\#SSCSP                             & \multicolumn{1}{c|}{65.00}                                  & \multicolumn{1}{c|}{56.79}                                  & \multicolumn{1}{c|}{54.06}                                  & \multicolumn{1}{c|}{95.63}                                  & \multicolumn{1}{c|}{74.69}                                  & \multicolumn{1}{c|}{79.06}                                  & \multicolumn{1}{c|}{80.00}                                  & \multicolumn{1}{c|}{87.81}                                  & 82.81          & 75.09(0.501)                                                                      \\ \hline
\rowcolor[HTML]{A6A6A6} 
\cellcolor[HTML]{9B9B9B}VA-EEGNet   & \multicolumn{1}{c|}{\cellcolor[HTML]{A6A6A6}77.50}          & \multicolumn{1}{c|}{\cellcolor[HTML]{A6A6A6}61.07}          & \multicolumn{1}{c|}{\cellcolor[HTML]{A6A6A6}63.12}          & \multicolumn{1}{c|}{\cellcolor[HTML]{A6A6A6}\textbf{98.43}} & \multicolumn{1}{c|}{\cellcolor[HTML]{A6A6A6}\textbf{96.56}} & \multicolumn{1}{c|}{\cellcolor[HTML]{A6A6A6}83.75}          & \multicolumn{1}{c|}{\cellcolor[HTML]{A6A6A6}84.37}          & \multicolumn{1}{c|}{\cellcolor[HTML]{A6A6A6}92.81}          & 88.43          & \cellcolor[HTML]{9B9B9B}82.93(0.658)                                              \\ \hline
\rowcolor[HTML]{D9D9D9} 
\cellcolor[HTML]{D9D9D9}VA-ConvNet  & \multicolumn{1}{c|}{\cellcolor[HTML]{D9D9D9}74.37}          & \multicolumn{1}{c|}{\cellcolor[HTML]{D9D9D9}56.07}          & \multicolumn{1}{c|}{\cellcolor[HTML]{D9D9D9}57.5}           & \multicolumn{1}{c|}{\cellcolor[HTML]{D9D9D9}97.5}           & \multicolumn{1}{c|}{\cellcolor[HTML]{D9D9D9}95.31}          & \multicolumn{1}{c|}{\cellcolor[HTML]{D9D9D9}82.18}          & \multicolumn{1}{c|}{\cellcolor[HTML]{D9D9D9}79.68}          & \multicolumn{1}{c|}{\cellcolor[HTML]{D9D9D9}87.5}           & 86.56          & \cellcolor[HTML]{D9D9D9}79.63(0.592)                                              \\ \hline
\rowcolor[HTML]{D9D9D9} 
\cellcolor[HTML]{D9D9D9}
\#DRDA      & \multicolumn{1}{c|}{\cellcolor[HTML]{D9D9D9}81.37}          & \multicolumn{1}{c|}{\cellcolor[HTML]{D9D9D9}\textbf{62.86}} & \multicolumn{1}{c|}{\cellcolor[HTML]{D9D9D9}\textbf{63.63}} & \multicolumn{1}{c|}{\cellcolor[HTML]{D9D9D9}95.94}          & \multicolumn{1}{c|}{\cellcolor[HTML]{D9D9D9}93.56}          & \multicolumn{1}{c|}{\cellcolor[HTML]{D9D9D9}88.19}          & \multicolumn{1}{c|}{\cellcolor[HTML]{D9D9D9}\textbf{85.00}} & \multicolumn{1}{c|}{\cellcolor[HTML]{D9D9D9}\textbf{95.25}} & \textbf{90.00} & \cellcolor[HTML]{D9D9D9}\textbf{83.98(0.679)}                                     \\ \hline
\rowcolor[HTML]{D9D9D9} 
\cellcolor[HTML]{D9D9D9}
*DA-ConvNet(ours)  & \multicolumn{1}{c|}{\cellcolor[HTML]{D9D9D9}\textbf{82.18}} & \multicolumn{1}{c|}{\cellcolor[HTML]{D9D9D9}60.00}          & \multicolumn{1}{c|}{\cellcolor[HTML]{D9D9D9}59.68}          & \multicolumn{1}{c|}{\cellcolor[HTML]{D9D9D9}98.12}          & \multicolumn{1}{c|}{\cellcolor[HTML]{D9D9D9}92.81}          & \multicolumn{1}{c|}{\cellcolor[HTML]{D9D9D9}\textbf{89.68}} & \multicolumn{1}{c|}{\cellcolor[HTML]{D9D9D9}84.06}          & \multicolumn{1}{c|}{\cellcolor[HTML]{D9D9D9}91.87}          & 88.43          & \cellcolor[HTML]{D9D9D9}82.98(0.659)                                              \\ \hline
\rowcolor[HTML]{A6A6A6} 
\cellcolor[HTML]{9B9B9B}
*DA-EEGNet(ours)   & \multicolumn{1}{c|}{\cellcolor[HTML]{A6A6A6}\textbf{85.00}} & \multicolumn{1}{c|}{\cellcolor[HTML]{A6A6A6}\textbf{66.78}} & \multicolumn{1}{c|}{\cellcolor[HTML]{A6A6A6}\textbf{73.43}} & \multicolumn{1}{c|}{\cellcolor[HTML]{A6A6A6}\textbf{98.75}} & \multicolumn{1}{c|}{\cellcolor[HTML]{A6A6A6}\textbf{95.93}} & \multicolumn{1}{c|}{\cellcolor[HTML]{A6A6A6}\textbf{90.62}} & \multicolumn{1}{c|}{\cellcolor[HTML]{A6A6A6}\textbf{88.75}} & \multicolumn{1}{c|}{\cellcolor[HTML]{A6A6A6}\textbf{96.25}} & \textbf{92.18} & \cellcolor[HTML]{9B9B9B}\textbf{87.52(0.750)}                                     \\ \hline
\end{tabular}

 \begin{tablenotes}
        \footnotesize
        \item[] White, light grey and dark shade indicate CSP-based methods, ConvNet-based method and EEGNet-based method, respectively. Methods with ”\#” adjust the data distribution among participants. Methods with “*” adjust the data distribution within the same participants
      \end{tablenotes}
    \end{threeparttable}
\end{table*}

\subsection{Experiment Results Analyses}
We conducted experiments on the aforementioned two datasets and compared our methods with several representative methods, particularly those based on transfer learning techniques.
The performance of different algorithms was first evaluated in the IIA dataset. The accuracy of each participant and the average accuracy of all participants (Kappa value) are shown in Table \ref{table3}. The EEGNet and ConvNet architectures with the SDDA framework were referred to as DA-EEGNet and DA-ConvNet, respectively. The two highest accuracy rates of each participant were displayed in bold to highlight experimental results. The items with white background in Table \ref{table3} were results from CSP-based algorithms. Light gray background rows were results from ConvNet which were generated using the same feature extractor and classifier network architecture as presented in \cite{schirrmeister2017deep}. Results from EEGNet and its corresponding domain adaptation method were marked with the dark gray background and referred to as EEGNet-based algorithms in the context. Algorithms with “\#” used EEG data from other participants in the dataset to help MI EEG classification of the target participant. And algorithms with “*” only exploited EEG data from the same participant recorded in different data acquisition sessions. Specifically, DRDA adjusted the distribution by generative adversarial network.

FBCSP is a classic MI-EEG decoding algorithm , and it was the winner of the competition for this dataset. Since the classification accuracy (average accuracy <60\%) of the CSP algorithm in this dataset is far lower than FBCSP, by adjusting the data distribution of different participants, CCSP and SSCSP achieve decoding capability similar to FBCSP. 
ConvNet and EEGNet related algorithms outperform FBCSP in classification accuracy at the expense of high computation cost. It is clear that compared to the vanilla network, DA-ConvNet and DA-EEGNet greatly improve accuracy in all participants, and DA-ConvNet achieveds the highest average accuracy. 
DADR is a representative work in the same training and testing semantics, which improved the classification performance by utilizing the useful information from other participants. The classification performance of DADR in this dataset exceeds a series of algorithms, such as C2CM(\cite{sakhavi2018learning}), MI-CNN(\cite{dose2018end}), SSMM(\cite{zheng2018sparse}). As can be seen from Table \ref{table3}, DA-EEGNet and DA-ConvNet surpass DADR in classification accuracy in most participants. We found that feature extraction layers for VA-ConvNet, DADR and DA-ConvNet had a high degree of similarity. However, DA-ConvNet has more advantages in classification performance than VA-ConvNet and DADR, demonstrating the effectiveness of SDDA framework in MI-EEG classification. 
Compared with VA-ConvNet, DADR has limited improvement in classification performance, which may be due to the large difference in data distribution among different participants, so that the feature extraction process is limited. By comparing the results from VA-EEGNet and DA-EEGNet in Table \ref{table3}, the SDDA framework greatly improves EEGNet's classification performance, indicating that the SDDA framework has no dependence on the network architecture.

To further investigate the effectiveness of the proposed SDDA framework, we evaluated the performance of the proposed algorithms in the IIB dataset. We used the same algorithms in Table \ref{table3} to demonstrate the generalization of different algorithms. The experimental results are shown in Table \ref{table4}. The decoding algorithms of manually extracted features FBCSP, SSCSP, and SSCSP show a relatively stable decoding ability in IIB dataset. Among them, the classification performance of FBCSP even surpasses VA-ConvNet. This shows that the classification performance of VA-ConvNet is limited with only three EEG channels. However, DADR, which has the similar feature extraction layers to VA-ConvNet, has better classification performance in IIB dataset than VA-ConvNet due to the introduction of information from other participants. 
By employing domain adaptation techniques, DA-ConvNet can enhance the classification performance of VA-ConvNet. However, given the limited spatial representation capacity of EEG signals captured by merely three channels, discriminating EEG signals poses a significant challenge. Consequently, incorporating EEG data from other participants with shared hidden information proves advantageous in augmenting the model generation process.
However, comparing VA-ConvNet, DADR, DA-ConvNet, VA-EEGNet and DA-EEGNet, we find that the architecture of the feature extraction network has a great influence on the classification accuracy. This is consistent with the conclusion in IIA. EEGNet related algorithms obtains better classification performance than ConvNet related algorithms in IIB, while the opposite performance is observed in IIA.
As illustrated in Table \ref{table:convnet}, the number of parameters in the ConvNet-based network is considerably greater than that in the EEGNet-based network. Given that only three recording channels capture a limited amount of essential motor imagery information, training a network with more parameters becomes challenging. However, despite the differences in network architecture, the classification performance is enhanced for nearly all participants with the assistance of the SDDA framework, which underscores its effectiveness.

\section{Analysis} \label{sec5}

In this section, we firstly performed effectiveness analysis on the main components from the proposed SDDA method to quantitively evaluate the contributions from each part of the framework. Then the convergence analysis based on the test results generated from the model training process was also presented to further investigate the robustness of the SDDA framework. Preliminary investigation on the distribution of high-dimensional features extracted by VA-EEGNet and DA-EEGNet in the two tested datasets were presented through feature visualization. At the end of this section, the sensitivity and generalizability were discussed by investigating the optimal trade-off hyperparameters for different participants to discuss the generalizability and stability of the proposed framework. 

\begin{figure*} 
\centering 
\includegraphics[width=0.9\textwidth]{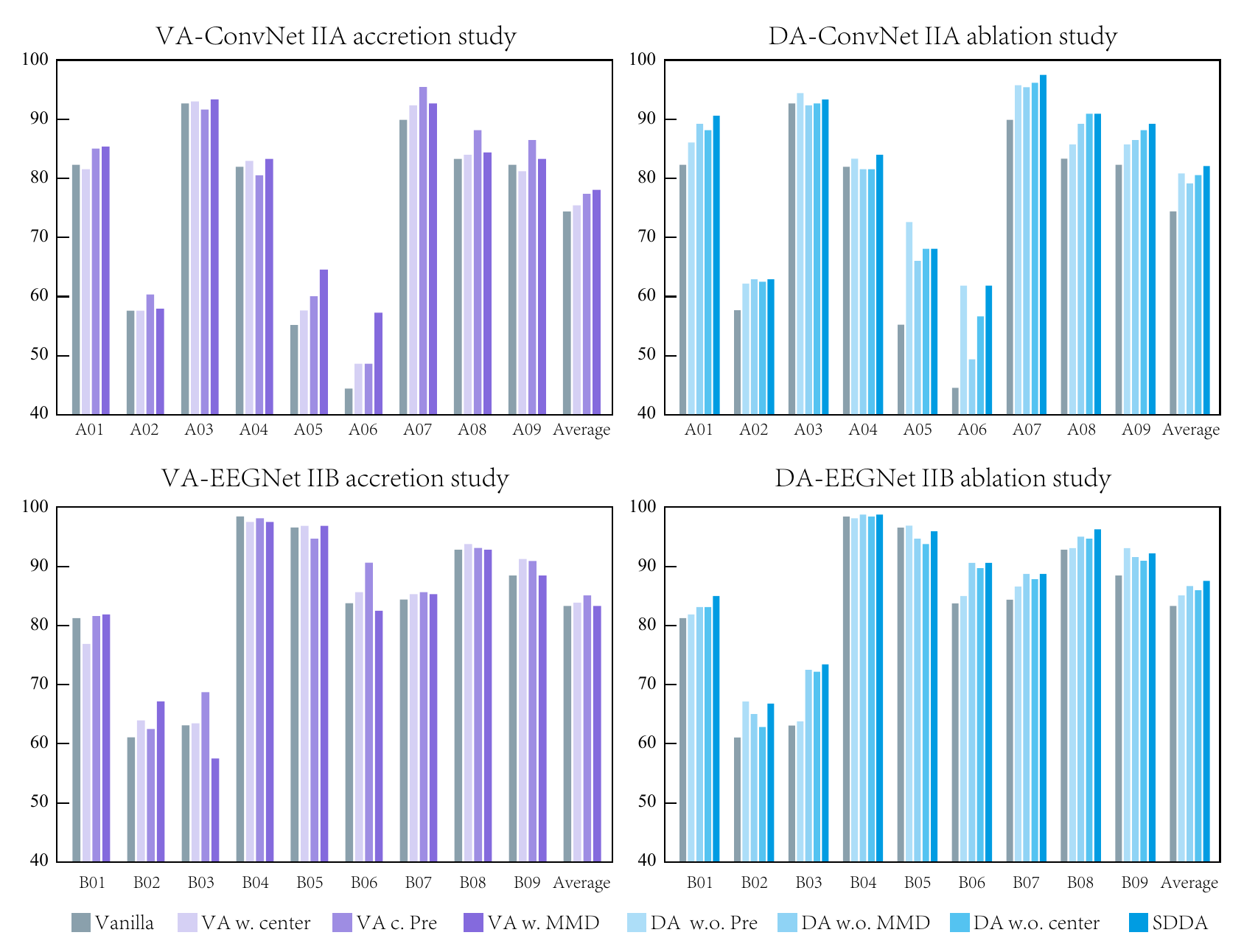} 
\caption{Summary of effective analysis results. VA c. Pre means vanilla network with proposed pre-processing methods. VA w. Center means vanilla network with cosine-based center loss. DA w.o. Center means SDDA framework without MMD loss. DA w.o. Center means SDDA framework without cosine-based center loss.}
\label{Fig.ablation} 
\end{figure*}

\subsection{Effectiveness Analysis}
Detailed analysis of the proposed SDDA framework was conducted to further investigate the influence of the pre-processing method, MMD loss, and cosine-based center loss on different participants in detail. All the experiments in this section were repeated five times and the average values were reported to minimize the influence from random results. In this sub-section, an accretion analysis of the pre-processing method, cosine-based center loss on vanilla networks were firstly conducted to evaluate the individual effect of the main components of the SDDA network. Then, ablation analysis of the SDDA framework was conducted by removing these components individually from the framework to evaluate the contribution of each component in the complete framework.

\subsubsection{Accretion Analysis on Vanilla Network}\mbox{}

In this subsection, we examined the impact of the three main components of the proposed SDDA by sequentially incorporating the designated preprocessing method, the center loss component, and the RHKS MMD loss into the vanilla network.

\paragraph{Preprocessing method for vanilla networks}\mbox{}

We replaced the original pre-processing method in vanilla networks with the processing method in the SDDA framework, which was referred to as VA c. Pre in Fig. \ref{Fig.ablation}. 
By introducing the proposed pre-processing method to the vanilla network, the average accuracy of ConvNet (VA-ConvNet c. Pre) reached 77.39\% in IIA dataset and 80.92\% in IIB dataset, with 2.98\% and 1.29\% increase in the accuracy compared to the VA-ConvNet, respectively. The average accuracy of EEGNet (VA-EEGNet c. Pre) reached 72.10\% in IIA and 85.1\% in IIB datasets with 3.16\% and 2.17\% increase in the accuracy compared to the VA-EEGNet, respectively. 

It can be seen from Fig. \ref{Fig.ablation}  that compared with the original pre-processing methods in the vanilla network, the pre-processing methods mentioned in this paper show better results in both datasets on most participants. The improvements are more prominent in participants with BCI illiteracy (A05, A06, B02). The pre-processing methods in the SDDA manage to effectively suppress noise and artifacts, and subsequently improve the signal quality, which is more effective for participants with poor BCI performance.

\paragraph{Cosine-based center loss for vanilla networks}\label{sec:center}\mbox{}

The effect of cosine-based center loss in vanilla networks (referred to as VA w. center in Fig. \ref{Fig.ablation}) was quantitively analyzed in this sub-section. For a fair comparison, the trade-off hyperparameter lambda in the cosine-based center loss was also searched from \{0, 0.2, 1, 2, 10, 15\} and the best classification results were reported for each participant. 

By including the cosine-base center loss into the loss function of the vanilla networks, the average accuracy of  ConvNet(VA-ConvNet w. center) reached 75.46\% in IIA dataset and 80.86\% in IIB dataset with 1.05\% and 1.23\% increase in the accuracy compared to the VA-ConvNet, respectively, while the average accuracy of EEGNet (VA-EEGNet w. center) reached 74.30\% and 83.84\% in IIA and IIB datasets with 5.36\% and 0.91\% increase in the accuracy compared to the VA-EEGNet, respectively.

As shown in Fig.\ref{Fig.ablation}, the cosine-based center loss is able to improve the performance of both vanilla networks. For participants with good results using the vanilla network (A01, A07, A08, A09, B06, B09), the improvement of these participants with cosine-based center loss is more effective. Differently from the pre-processing method, cosine-based center loss works effectively as a strong regularization method, which constrains the impact of the intra-class feature discrepancies in a limited range, and thus suppresses the outliers in both domains. Such characteristic of cosine-based center loss can work complementarily with the pre-processing method in the previous section.

It is worth noticing that, when using Euclidean distance as center loss instead of the cosine-based center loss, the model convergence and classification performance are significantly lower compared to the vanilla model according to our preliminary results (\textgreater20\% decrease in accuracy, some models even failed to converge). Therefore, results with center loss using Euclidean distance are not included in this paper. This further illustrates the superiority of the cosine-based center loss proposed in this paper. A proper representation is really important for MI-EEG classification when using domain adaptation technics.

\paragraph{MMD loss for vanilla networks}\mbox{}

By adopting Siamese structure in the vanilla network, the features were mapped into the RHKS and regularized with a MMD loss, which was referred as VA w. MMD in the Fig. \ref{Fig.ablation}.  The average accuracy of ConvNet (VA-ConvNet w. MMD) reached 78.04\% in IIA and 80.95\% in IIB, with 3.63\% and 1.32\% increase in the accuracy compared to the VA-ConvNet, respectively. The average accuracy of EEGNet (VA-EEGNet w. MMD) reached 73.98\% in IIA and 83.32\% in IIB with 5.04\% and 0.38\% increase in the accuracy compared to the VA-EEGNet, respectively.

The results are summarized in Fig. \ref{Fig.ablation}, which suggest the improvements are more prominent in participants with BCI illiteracy (A05, A06, B02). Combining the results from VA w. MMD and VA c. Pre, it is interesting to see that, despite the alignment on different level (raw data distribution vs embedding features distribution), participants with poor BCI performance can always benefit from the alignment process. These results strongly demonstrate that the domain adaptation methods for MI EEG classification are effective.

\subsubsection{Ablation Analysis on SDDA Framework}\mbox{}
Ablation analysis was performed for further investigation of the three main components in the SDDA framework. The pre-processing method, cosine-based center loss and the RHKS MMD loss were removed in turn to quantitatively investigate the contribution of each main component in the proposed SDDA framework.

\paragraph{Preprocessing method in SDDA}\mbox{}

In this part of the ablation analysis, the pre-processing method in SDDA framework was substituted by the pre-processing method mentioned in ConvNet (\cite{schirrmeister2017deep}) (referred to as DA w.o. Pre in Fig \ref{Fig.ablation}). The average accuracy of ConvNet in IIA dataset reached 80.86\% and 81.44\% in the IIB dataset, with 1.22\% and 1.54\% decrease in the accuracy compared to the ConvNet-based SDDA. The average accuracy of EEGNet in IIA reached 76.08\% and 85.06\% in IIB dataset, with 3.54\% and 2.46\% decrease in the accuracy compared to the EEGNet-based SDDA.

As shown in Fig. \ref{Fig.ablation}, pre-processing method in SDDA poses a more prominent impact in models generated from IIB dataset than that from IIA. Given that there are only three recording channels in the IIB dataset, the signal-to-noise ratio of the data in IIB is remarkably lower than the data in IIA. The results demonstrate the pre-processing method in SDDA indeed constructs useful features which improve MI classification.

\paragraph{Cosine-based center loss in SDDA}\mbox{}

To quantify the effect of cosine-based center loss in SDDA, we set $\lambda_1$ in Equation (\ref{eq1}) to zero and searched $\lambda_2$ from \{0, 0.02, 0.05, 0.1, 0.2, 0.5\}. The best classification results were presented for each participant in Fig. \ref{Fig.ablation}, which were referred to as DA w.o. center. By removing the cosine-based center loss in the SDDA framework, the average accuracy of ConvNet reached 80.55\% in IIA and 82.45\% in IIB dataset, with 1.51\% and 0.53\% decrease in the accuracy compared to the complete SDDA framework. The average accuracy of EEGNet reached 76.30\% in IIA and 85.94\% in IIB, respectively, which decreased by 3.36\% in IIA and 1.58\% in IIB compared to the EEGNet-based SSDA framework in the accuracy.

It can be seen from Fig. \ref{Fig.ablation} that after removing the cosine-based center loss, the accuracy of almost every participant is lower than that of SDDA, which shows that it is reasonable to use the cosine-based center loss as a regularization term in the training process. 

\paragraph{MMD loss in SDDA}\mbox{}

In this part of the ablation analysis, MMD loss was removed from the SDDA framework, $\lambda_2$ in Equation (\ref{eq1}) was set to 0 and $\lambda_1$ was searched  from \{0, 0.2, 1, 2, 10, 15\}. The highest classification results were presented for each participant in Fig. \ref{Fig.ablation}, which were noted as DA w.o. MMD. As shown in Fig. \ref{Fig.ablation}, the average accuracy of ConvNet (VA-ConvNet Pre+center) in the IIA dataset reached 79.16\% and 82.28\% in the IIB dataset, with 2.84\% and 0.70\% decrease in the accuracy, compared to the complete SDDA framework respectively. While the average accuracy of EEGNet (VA-EEGNet Pre+center) reached 77.27\% in IIA and 86.66\% in IIB dataset, with 2.39\% and 0.86\% decrease in the accuracy, compared to the the complete SDDA framework respectively. Different from the results generated by only adding cosine-based center loss in Section \ref{sec:center}, the accuracy of VA Pre+center in most participants in both datasets gained major improvements, especially for the participants in IIB dataset. The classification accuracy for all participants in these two datasets was improved with such configuration, which demonstrated the validity of domain adaptation for adjusting MI cross-session classification

In summary, the complete SDDA exhibited great efficacy and integrity as a universal plug-and play framework compatible for various artificial neural networks. The experimental results in this section were consistent with Theorem \ref{theorem2}, which demonstrated the necessity of minimizing error in the source domain as well as distribution discrepancy between source and target domain, in order to achieve good classification performance in both source and target domain. This is also consistent with the architecture and effective analysis results of SDDA framework, which further supports the rationale of integrating domain adaptation technologies to solve the cross-session MI-EEG classification problem. 

\subsection{Model training profile}
In this subsection, we recorded the fluctuation of various neural network models during the training process, using testing accuracy as the metric.
Two participants, A07 and B01, one from each dataset, were selected to showcase the difference in the training process between the SDDA framework and vanilla networks, as they demonstrated the most significant improvement in classification performance compared to the vanilla networks. All models were set to run 500 epochs in the first training stage and 300 epochs in the second training stage. 

The training profiles of the proposed SDDA framework are depicted in Fig. \ref{Fig.training} to illustrate the stability of the proposed method during training. As seen in Fig. \ref{Fig.training}, the SDDA framework displays reduced fluctuations in test accuracy as training epochs increase compared to vanilla neural networks throughout the model generation process. These findings align with the earlier results from the effectiveness analysis, which suggest that our proposed SDDA framework has less training fluctuation and enhance the vanilla networks' resilience against noise and outliers, even during the model generation process. Interestingly, when incorporating additional training data into the last 300 epochs, the test accuracy increases immediately. DA-ConvNet appears to benefit significantly from this process, while DA-EEGNet is less sensitive to the extra data. This observation is also consistent with our results from the effectiveness analysis, where ConvNet, with a higher number of parameters, was more "data-hungry" compared to EEGNet. SDDA introduces appropriate regularization during training, compelling the vanilla network to learn more valuable information with the same number of training samples. Furthermore, as seen in Fig. \ref{Fig.training}, the convergence curve of SDDA exhibits fewer fluctuations than the vanilla network, indicating enhanced robustness against noise.

\begin{figure}
\centering 
\includegraphics[width=0.48\textwidth]{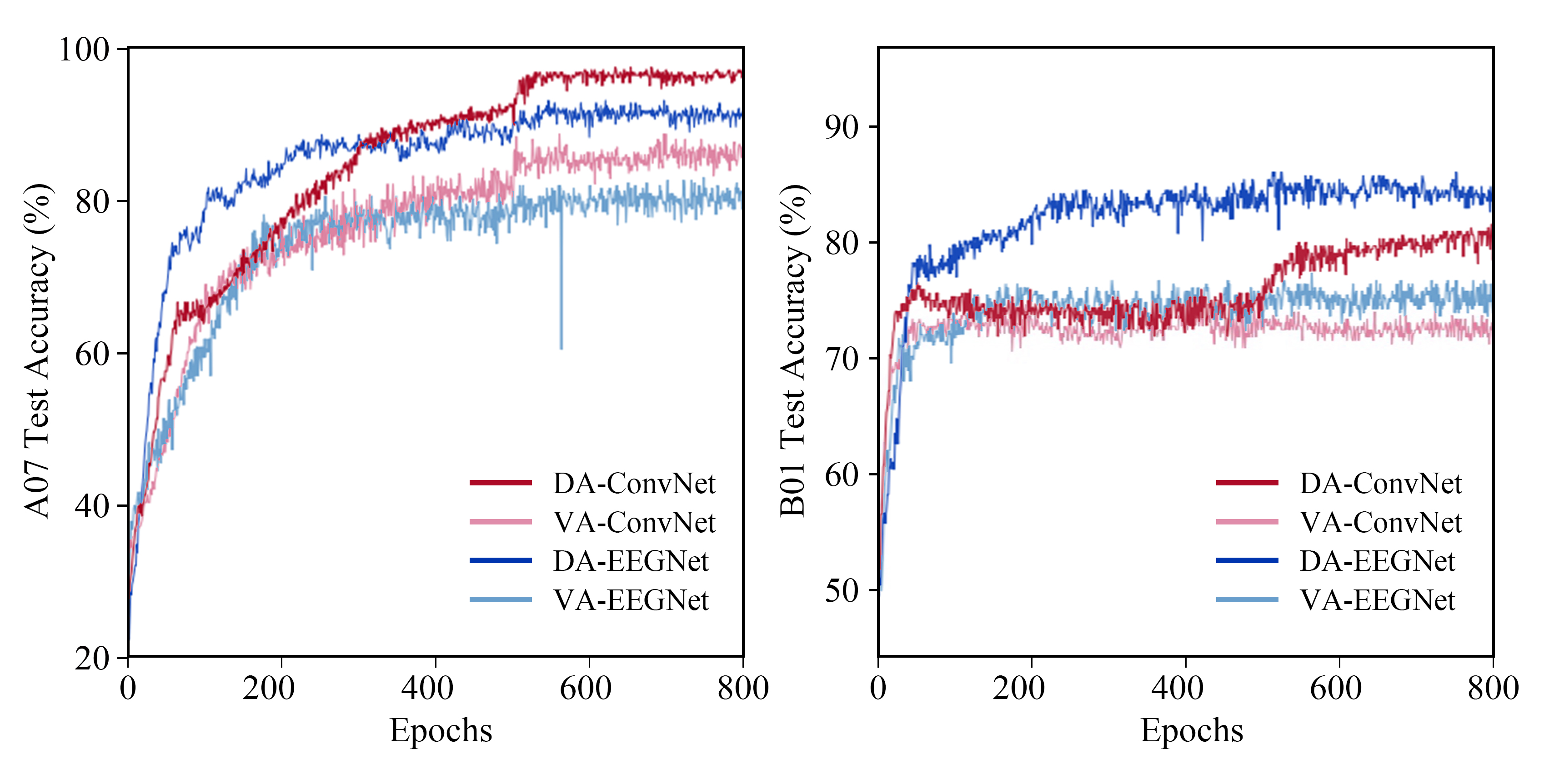} 
\caption{Relationship between training epochs and testing accuracy of different models of IIA and IIB datasets. DA refers to networks which were trained with SDDA framework. VA refers to networks which were trained with vanilla network.}
\label{Fig.training} 
\end{figure}

\subsection{Feature Visualization}
T-SNE (\cite{VanderMaaten2008}) was further utilized to investigate the embedding features of both vanilla networks and the proposed SDDA framework. Embedding features from DA-EEGNet were visualized from two different participants (A03, B06), whose model performance improved the most under EEGNet-based SDDA framework. 

The visualization results are shown in Fig. \ref{Fig.tsne}. According to the visualization results, the feature distribution from SDDA is more discriminative than that of the vanilla network, which is consistent with the results presented in Table \ref{table3} and Table \ref{table4}. Generally, higher classification accuracy is associated with more separable feature distributions after t-SNE dimensionality reduction.
\begin{figure}
\centering 
\includegraphics[width=0.48\textwidth]{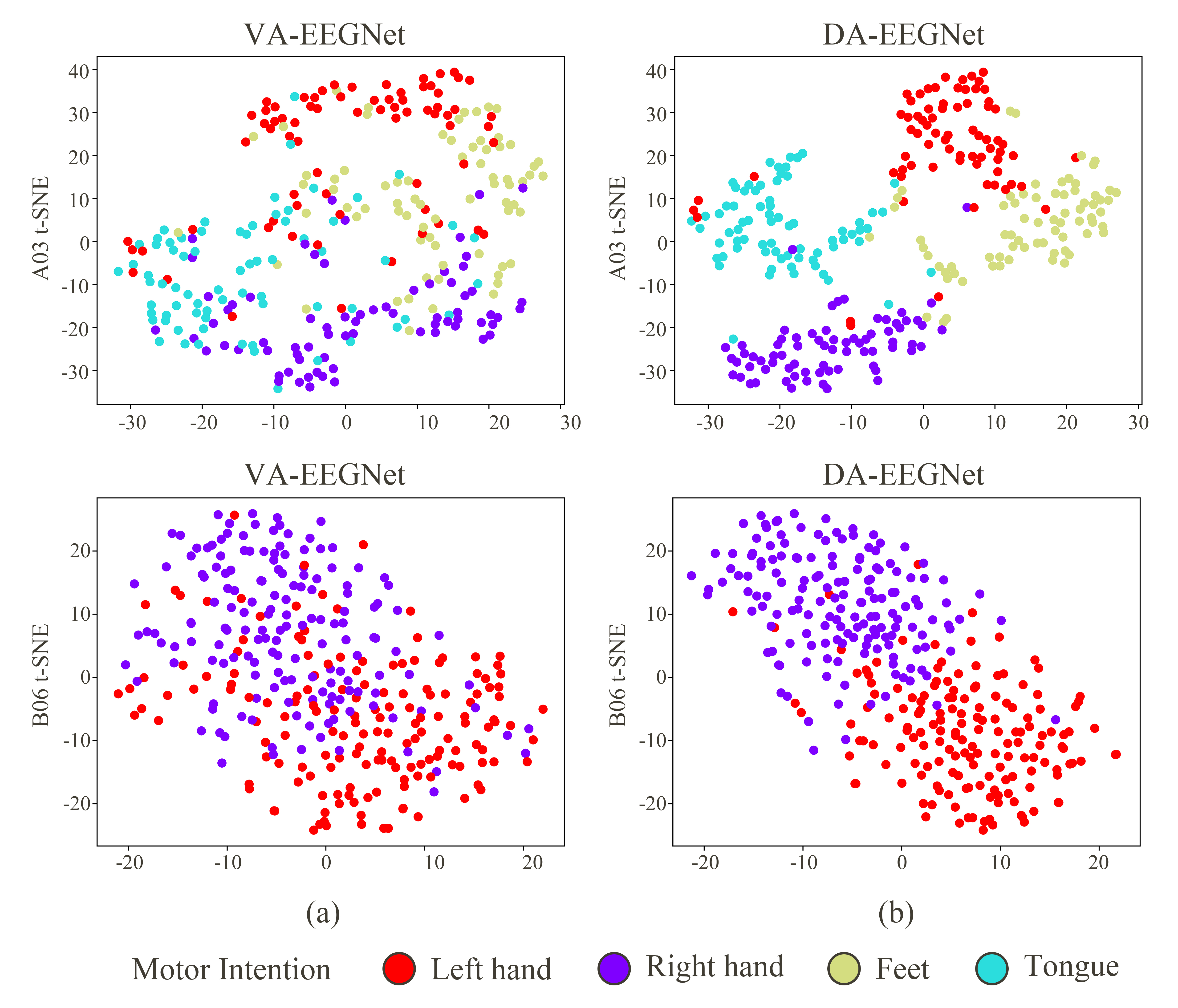} 
\caption{Visualization of embedding feature by t-SNE. (a) Left column shows the feature distribution of vanilla EEGNet. (b) Right column shows the feature distribution of SDDA with EEGNet.}
\label{Fig.tsne} 
\end{figure}
\subsection{Trade-off Parameters Sensitivity}

The loss function of the proposed SDDA framework in equation (\ref{eq1}) consists of three terms, $\mathcal{L}_s$ , $\mathcal{L}_c$, and $\mathcal{L}_d$. Different weights of different losses will influence the classification results. As stated in equation (\ref{eq1}), larger $\lambda_1$ indicates a greater contribution of the cosine-based center loss to the total loss, while the larger $\lambda_2$ indicates greater contribution of MMD loss to the total loss, and vice versa. For example, when all participants use $\lambda_1=10, \lambda_2=0.01$ as trade-off parameters, the optimal average accuracy of DA-ConvNet and DA-EEGNet in IIA reached 79.55\% and 74.84\%, respectively. When  $\lambda_1=15, \lambda_2=0.01$ the optimal average accuracy of DA-ConvNet and DA-EEGNet in IIB reach 81.78\% and 86.11\%. The weight of cosine-based center loss is not as sensitive as that of the MMD loss, as the cosine function constrained the function outputs between [0, 1]. These results also confirm that the cosine-based center loss itself has very good robustness in MI EEG classification.

A relatively small list of trade-off parameters ($6\times6$ matrix) was grid-searched and optimal trade-off parameters of different participants from both datasets were presented.

As shown in Fig. \ref{Fig.parameter}, optimal trade-off parameters are slightly different. Among them, the trade-off parameters in the IIB dataset show less centralized distribution compared to that of IIA, which indirectly indicates that the data distribution of different participants in this dataset is quite different. In addition, it is interesting to see that participants with high MI-EEG performance share similar trade-off parameters (especially $\lambda_1=0.01, \lambda_2=0.2 \text{ or } 1  \text{ or } 10  \text{ or } 15$). These results are clear evidence for our prior hypothesis about negative transfer on the cross-participant MI classification that use all participants’ data in the model generation. Participants with large data distribution discrepancy (far away from the clusters in Fig. \ref{Fig.parameter}.) will cause negative transfer and pose negative effects on model performance. This might also be the reason why our SDDA framework demonstrates better performance than the GAN-based methods in the literature, even though our SDDA framework uses much less data in the model training process.
\begin{figure}
\centering 
\includegraphics[width=0.48\textwidth]{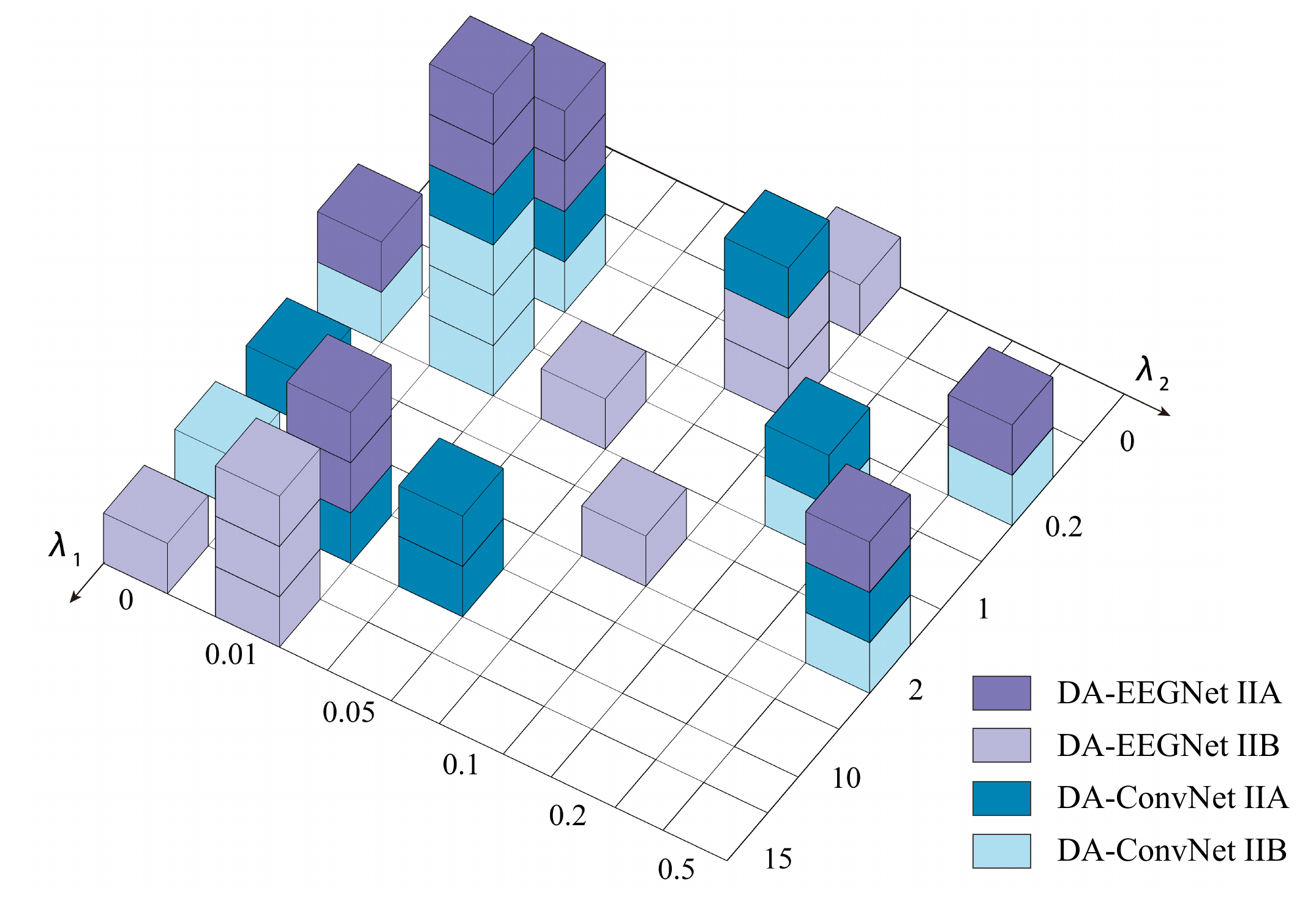} 
\caption{Best Trade-off parameters of SDDA for different participants. $\lambda_1$ and $\lambda_2$ represent the weight of the cosine-based center loss and MMD loss, respectively. Each block indicates one set of optimal trade-off parameter for one participant in the corresponding dataset. }
\label{Fig.parameter} 
\end{figure}

\section{Benefits and limitations} \label{sec6}
\subsection{Benefits}
The proposed SDDA framework offers three advantages, including:
\begin{enumerate}
    \item The SDDA framework is a transferable deep domain adaptation framework that can be applied to any convolutional neural network suitable for decoding motor imagery, thereby enhancing the classification performance of the vanilla network.
    \item By preprocessing and constructing loss functions, the SDDA framework addresses the problem of motor imagery signal drifting over time. It accomplishes this without introducing additional parameters, making it friendly to EEG signals with a small amount of data.
    \item The SDDA framework improves the decoding accuracy of the target participant without relying on the data of other participants, safeguarding the rights of EEG participants.
\end{enumerate}

\subsection{limitations}
It is noted that the proposed SDDA framework has the following limitations.
\begin{enumerate}
    \item The SDDA framework is a generic network architecture that only enhances the performance of the vanilla network, and the classification results depend on the design of the vanilla network.
    \item In a real online system, the EEG data of the target domain may be generated sequentially, which presents a challenge to the measurement of the target domain distribution.
    \item The hyperparameters $\lambda_1$ and $\lambda_2$ may need to be readjusted when decoding data from other MI datasets.
\end{enumerate}

\section{Conclusion} \label{sec7}
In this paper, we presented an SDDA framework for cross-session MI classification, designed to seamlessly integrate with most existing ANNs without necessitating alterations to the original network architecture. We incorporated a novel pre-processing method aimed at constructing domain invariants based on prior knowledge. Embedding features for both source and target domains were aligned in the RKHS using MMD loss. Additionally, a cosine-based center loss was included in the framework to enhance the generalizability and robustness of the classifiers.
The proposed deep domain adaptation framework was rigorously validated on two publicly available MI-EEG datasets using two classic and widely-employed EEG classification CNN architectures. Our SDDA framework not only substantially improved the performance of the vanilla network but also surpassed representative transfer learning-based methods.

Furthermore, the SDDA framework lays the groundwork for addressing cross-session MI classification challenges by offering practical neural network solutions that harness domain adaptation technologies. As cross-session variability is a pervasive issue in motor imagery, the SDDA framework can be applied to any motor imagery scenario that requires participant pretraining. Moreover, the SDDA framework plays a vital role in enhancing the motor imagery accuracy of specific participants. By setting and adjusting individual-specific trade-off parameters, $\lambda_1$ and $\lambda_2$, the SDDA framework can improve the motor imagery classification accuracy of specific participants without modifying the vanilla network structure. This demonstrates the potential of the SDDA framework to play a significant role in brain-computer interface applications tailored to individual training.

\section*{Appendix}

The SDDA framework employs Siamese networks, which consume more computational resources compared to vanilla networks. To further investigate the impact of computational resources on the experimental results, we ensured a fair comparison between computational resource effects and the SDDA framework by consistently adopting the preprocessing method proposed in this paper. This maintains the consistency of EEG data when inputting into the neural network model. Table \ref{table:batchsize} displays the classification performance of EEGNet-based models on the BCIIV-IIB dataset under different batch sizes. The results indicate that varying batch sizes do indeed have an impact on the experimental outcomes. However, as shown in Table \ref{table:batchsize}, the classification performance of the DA-EEGNet surpasses that of all VA-EEGNet models in the majority of participants. This suggests that although the SDDA framework increases computational resources to some extent, the additional computational overhead is justifiable, as the improvement in classification performance is significant.

\begin{table*}[!hbt]
\caption{Impact of Computational Resources on the Results of BCIIV-IIB Dataset: An Illustration with EEGNet}
\label{table:batchsize}
\begin{tabular}{|c|ccccccccc|c|}
\hline
\multirow{2}{*}{Method} & \multicolumn{9}{c|}{Paiticipant ID}                                                                                                                                                                                                                                                                                            & \multirow{2}{*}{\begin{tabular}[c]{@{}c@{}}Average acc\\ (kappa)\end{tabular}} \\ \cline{2-10}
                        & \multicolumn{1}{c|}{B01}            & \multicolumn{1}{c|}{B02}            & \multicolumn{1}{c|}{B03}            & \multicolumn{1}{c|}{B04}            & \multicolumn{1}{c|}{B05}            & \multicolumn{1}{c|}{B06}            & \multicolumn{1}{c|}{B07}            & \multicolumn{1}{c|}{B08}            & B09            &                                                                                \\ \hline
VA-EEGNet c.Pre(all)   & \multicolumn{1}{c|}{83.12}          & \multicolumn{1}{c|}{65.71}          & \multicolumn{1}{c|}{71.87}          & \multicolumn{1}{c|}{97.81}          & \multicolumn{1}{c|}{94.06}          & \multicolumn{1}{c|}{86.25}          & \multicolumn{1}{c|}{87.50}          & \multicolumn{1}{c|}{95.00}          & 88.75          & 85.56(0.711)                                                                   \\ \hline
VA-EEGNet c.Pre(128)   & \multicolumn{1}{c|}{82.50}          & \multicolumn{1}{c|}{64.28}          & \multicolumn{1}{c|}{71.56}          & \multicolumn{1}{c|}{98.12}          & \multicolumn{1}{c|}{94.68}          & \multicolumn{1}{c|}{89.68}          & \multicolumn{1}{c|}{87.18}          & \multicolumn{1}{c|}{92.81}          & 91.25          & 85.78(0.716)                                                                   \\ \hline
VA-EEGNet c.Pre(64)    & \multicolumn{1}{c|}{80.93}          & \multicolumn{1}{c|}{\textbf{66.78}} & \multicolumn{1}{c|}{74.06}          & \multicolumn{1}{c|}{97.81}          & \multicolumn{1}{c|}{94.37}          & \multicolumn{1}{c|}{86.87}          & \multicolumn{1}{c|}{88.12}          & \multicolumn{1}{c|}{94.06}          & 90.62          & 85.96(0.719)                                                                   \\ \hline
VA-EEGNet c.Pre(32)    & \multicolumn{1}{c|}{82.50}          & \multicolumn{1}{c|}{65.00}          & \multicolumn{1}{c|}{\textbf{75.31}} & \multicolumn{1}{c|}{97.81}          & \multicolumn{1}{c|}{94.06}          & \multicolumn{1}{c|}{85.00}          & \multicolumn{1}{c|}{85.93}          & \multicolumn{1}{c|}{94.37}          & 91.25          & 85.69(0.714)                                                                   \\ \hline
VA-EEGNet c.Pre(16)     & \multicolumn{1}{c|}{81.56}          & \multicolumn{1}{c|}{62.50}          & \multicolumn{1}{c|}{68.75}          & \multicolumn{1}{c|}{98.12}          & \multicolumn{1}{c|}{94.68}          & \multicolumn{1}{c|}{\textbf{90.62}} & \multicolumn{1}{c|}{85.62}          & \multicolumn{1}{c|}{93.12}          & 90.93          & 85.10(0.702)                                                                   \\ \hline
DA-EEGNet(16)          & \multicolumn{1}{c|}{\textbf{85.00}} & \multicolumn{1}{c|}{\textbf{66.78}} & \multicolumn{1}{c|}{73.43}          & \multicolumn{1}{c|}{\textbf{98.75}} & \multicolumn{1}{c|}{\textbf{95.93}} & \multicolumn{1}{c|}{\textbf{90.62}} & \multicolumn{1}{c|}{\textbf{88.75}} & \multicolumn{1}{c|}{\textbf{96.25}} & \textbf{92.18} & \textbf{87.52(0.750)}                                                          \\ \hline
\end{tabular}
\begin{tablenotes}
        \footnotesize
        \item In the "Method" column, the number in parentheses represents the batch size, "all" means the batch size is the number of trials in the training set.
        \item "VA-EEGNet c.Pre" denotes the vanilla EEGNet model with the preprocessing method proposed in this paper. 
      \end{tablenotes}
\end{table*}

\section*{Acknowledgments}
This work was supported in part by the National Natural Science Foundation of China under Grant 82102174, National Key Research and Development Program of China under Grant 2022ZD0208900, China Postdoctoral Science Foundation under Grant 2021TQ0243, and Tianjin Science and Technology Planning Project under Grant 20JCQNJC01250.

\newpage
\bibliographystyle{cas-model2-names}

\bibliography{SDDA}

\end{document}